\documentclass[letterpaper, 10 pt, conference]{ieeeconf}  

\IEEEoverridecommandlockouts                              
\usepackage{authblk}
\usepackage{balance}
\usepackage{cite}
\usepackage{balance}
\usepackage{cite}
\usepackage{amsmath,amssymb,amsfonts}
\usepackage{algorithmic}
\usepackage{graphicx}
\usepackage{textcomp}
\usepackage{todonotes}
\usepackage{cleveref}
\usepackage{url}
\usepackage[hang,flushmargin]{footmisc}

\newcommand{\pseudocode}{\texttt}

\begin{document}

\title{\LARGE \bf
Federated Learning for Coalition Operations*
}

\author{D. Verma$^{1}$, S. Calo$^{1}$, S. Witherspoon $^{1}$, E. Bertino,${^2}$ A. Abu Jabal${^2}$
\\
A. Swami$^{3}$, G. Cirincione$^{3}$, S. Julier$^{4}$, G. White${^5}$, G. de Mel${^5}$, G. Pearson $^{6}$%
\\
\\$^{1}$IBM T. J. Watson Research Center, Yorktown Heights, NY 10598, USA %
\\$^{2}$Computer Science Dept, Purdue University,  West Lafayette, IN-47907, USA %
\\$^{3}$ U.S. Army Research Lab, 2800 Powder Mill Road, Adelphi, MD 20783, U.S.A.%
\\$^{4}$Computer Science Dept, University College London, 66 Gower St, London, WC1E6BT,U.K. %
\\$^{5}$IBM Research UK, Hursley Park, Winchester, SO21 2JN, U.K. %
\\$^{6}$Defence Science \& Technology Laboratory, Porton Down, Salisbury SP4 0JQ, U.K %
\\
\thanks{This research was sponsored by the U.S. Army Research Laboratory and the U.K. Ministry of Defence under Agreement Number W911NF-16-3-0001. The views and conclusions contained in this document are those of the authors and should not be interpreted as representing the official policies, either expressed or implied, of the U.S. Army Research Laboratory, the U.S. Government, the U.K. Ministry of Defence or the U.K. Government. The U.S. and U.K. Governments are authorized to reproduce and distribute reprints for Government purposes notwithstanding any copyright notation hereon.}
}

\maketitle
\thispagestyle{empty}
\pagestyle{empty}


\label{sec:abstr}

\begin{abstract}
Machine Learning in coalition settings requires combining insights available from data assets and knowledge repositories distributed across multiple coalition partners. In tactical environments, this requires sharing the assets, knowledge and models in a bandwidth-constrained environment, while staying in conformance with the privacy, security and other applicable policies for each coalition member. Federated Machine Learning provides an approach for such sharing. In its simplest version, federated machine learning could exchange training data available among the different coalition members, with each partner deciding which part of the training data from other partners to accept based on the quality and value of the offered data. In a more sophisticated version, coalition partners may exchange models learnt locally, which need to be transformed, accepted in entirety or in part based on the quality and value offered by each model, and fused together into an integrated model.  In this paper, we examine the challenges present in creating federated learning solutions in coalition settings, and present the different flavors of federated learning that we have created as part of our research in the DAIS ITA. The challenges addressed include dealing with varying quality of data and models, determining the value offered by the data/model of each coalition partner, addressing the heterogeneity in data representation, labeling and AI model architecture selected by different coalition members, and handling the varying levels of trust present among members of the coalition. We also identify some open problems that remain to be addressed to create a viable solution for federated learning in coalition environments. 
 
\end{abstract}
\section{Introduction}
\label{sec:intro}

Artificial Intelligence and specifically Machine Learning based approaches can enhance the military-technical advantage of United States and its allies in many aspects of their joint coalition operations. However, the application of Artificial Intelligence/Machine Learning (AI/ML) must overcome significant technical challenges, especially when used at the tactical edge for coalition operations such as the operations described in~\cite{white2019dais-ita}. In this paper, we look at the challenges involved in the application of AI/ML in the context of coalition operations, and discuss advances in federated learning which allow us to apply these techniques to coalition operations. 


We begin this paper with a brief overview of AI/ML in a generic context, followed by an overview of coalition operations and the two primary types of machine learning scenarios that emerge within coalition operations. Then we consider the challenges involved in each of these two scenarios, and show how generating policies addresses the challenges inherent in both of those scenarios.

\section{A brief overview of AI/ML}
\label{sec:ai}

The goal of applying AI/ML in coalition operations is the improvement of  specific tasks that arise in the context of coalition operations. This requires a systematic approach to assess what coalition tasks are appropriate to be AI-enabled, the level of autonomy that is desired, and how AI/ML can overcome resource constraints (such as: low bandwidths; low CPU power; or running on batteries) and environmental uncertainties in the deployed environment (such as: weather conditions; adversarial actions; or civilian life).

\begin{figure}
      \centering
      \includegraphics[width=0.9\columnwidth]{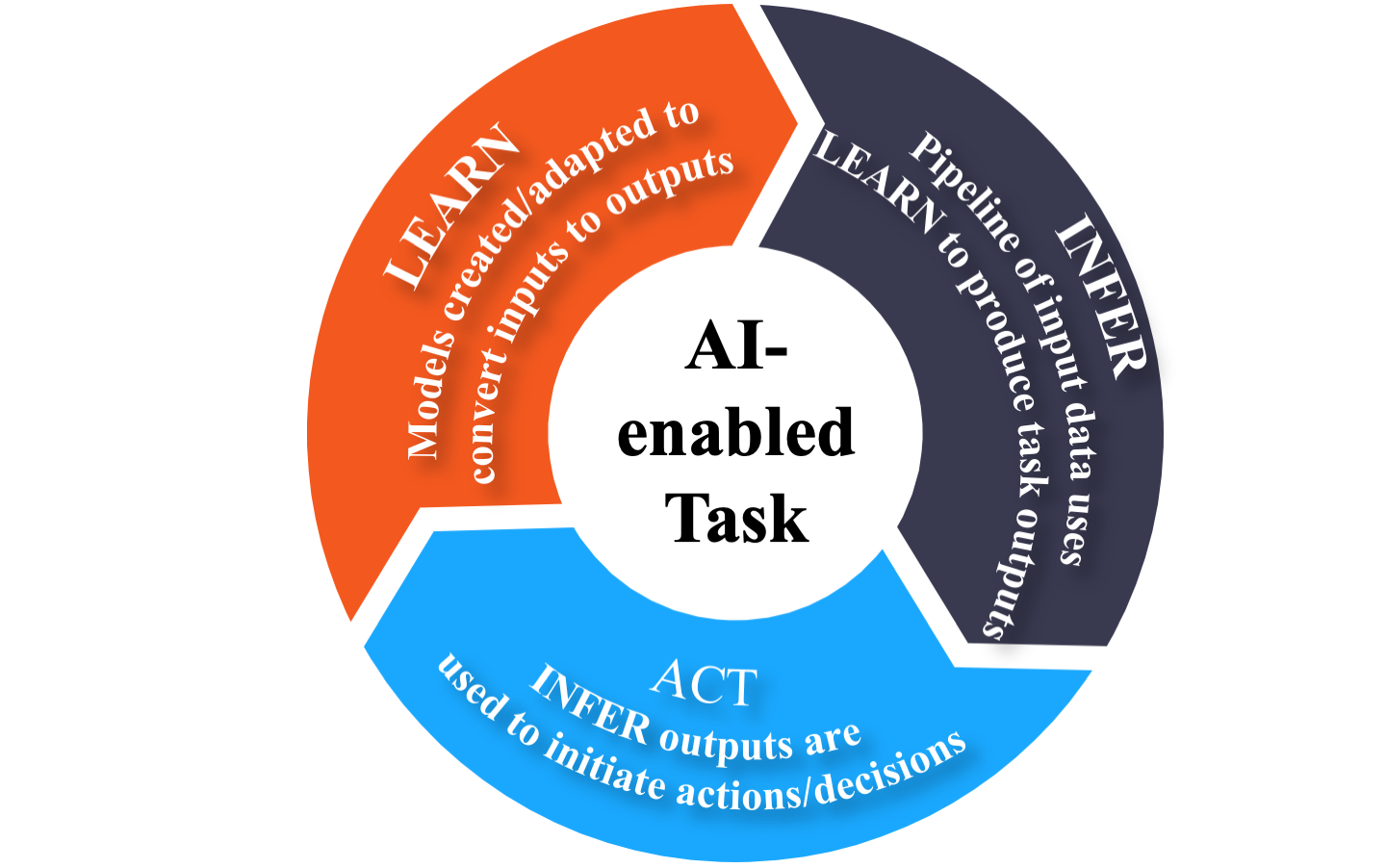}
      \caption{The Learn-Infer-Act Cycle }
      \label{fig:fig1}
\end{figure}

As described in \cite{cirin2019}, the process for creating, deploying and using AI/ML can be viewed as occurring in three phases: Learning, Inferring and Acting.  The process and its key design inputs are described below and shown in Figure~\ref{fig:fig1}. 

\textbf{Learn Phase.} In the Learn phase, one or more models are created or adapted to convert inputs into outputs as required by the AI enabled task. This phase uses training data to create a model, and in addition to collecting and curating the data, may include performance evaluation, validation of model accuracy, creating auxiliary explainability approaches for the model, and incremental adaptation of the model as more training data are obtained. 

\textbf{Infer Phase:} In the Infer phase, the model or models created in the learn phase are used in a pipeline of activities on input data to produce an output required by the coalition task. 

\textbf{Act Phase:} In the Act phase, outputs of Infer phase are used to take an action or feed a decision. The type of action/decision can be varied, and may involve input from humans/warfighters and other machines. 

The three phases usually happen in a cycle, since the results of an action can be used to improve the model in an iterative manner. 

There are many types of model architectures, training algorithms and distribution architectures for learning and inference that we will not describe in this paper, but refer the reader to \cite{cirin2019} for a more detailed treatment of the issues involved in multi-domain and coalition operations. 

In coalition contexts, different partners of the coalitions may choose to cooperate and collaborate at different phases of the three-phase cycle. 

\textbf{Federation during Learn Phase:} The coalition partners may decide to collaborate during the learning phase in order to create a joint model that all the partners could share. This federation is the main focus of this paper, and is described in more detail in the subsequent sections. 

\textbf{Federation during Infer Phase:} The coalition partners may share the inferences they reach using their individual models. The models used could be different or the same depending on whether the partners elected to coordinate during the learn phase. Even if the partners use the same model, they may reach different outputs during the infer phase if they have different inputs, or view the same input with different modalities. 

\textbf{Federation during Act Phase:} The coalition partners may decide to coordinate their actions to be performed based on their individual or federated inference. Coordination of the action, and informing other partners of actions they may take can improve the effectiveness of the coalition task. 
\section{Coalition Learning Scenarios}
\label{sec:scenario}
In military coalitions where partners are collaborating and coordinating to create a common model, they may perform this collaboration in one of two modes: \emph{data sharing mode}; and \emph{model sharing mode}. 

\begin{figure}[h]
      \centering
      \includegraphics[width=0.8\columnwidth]{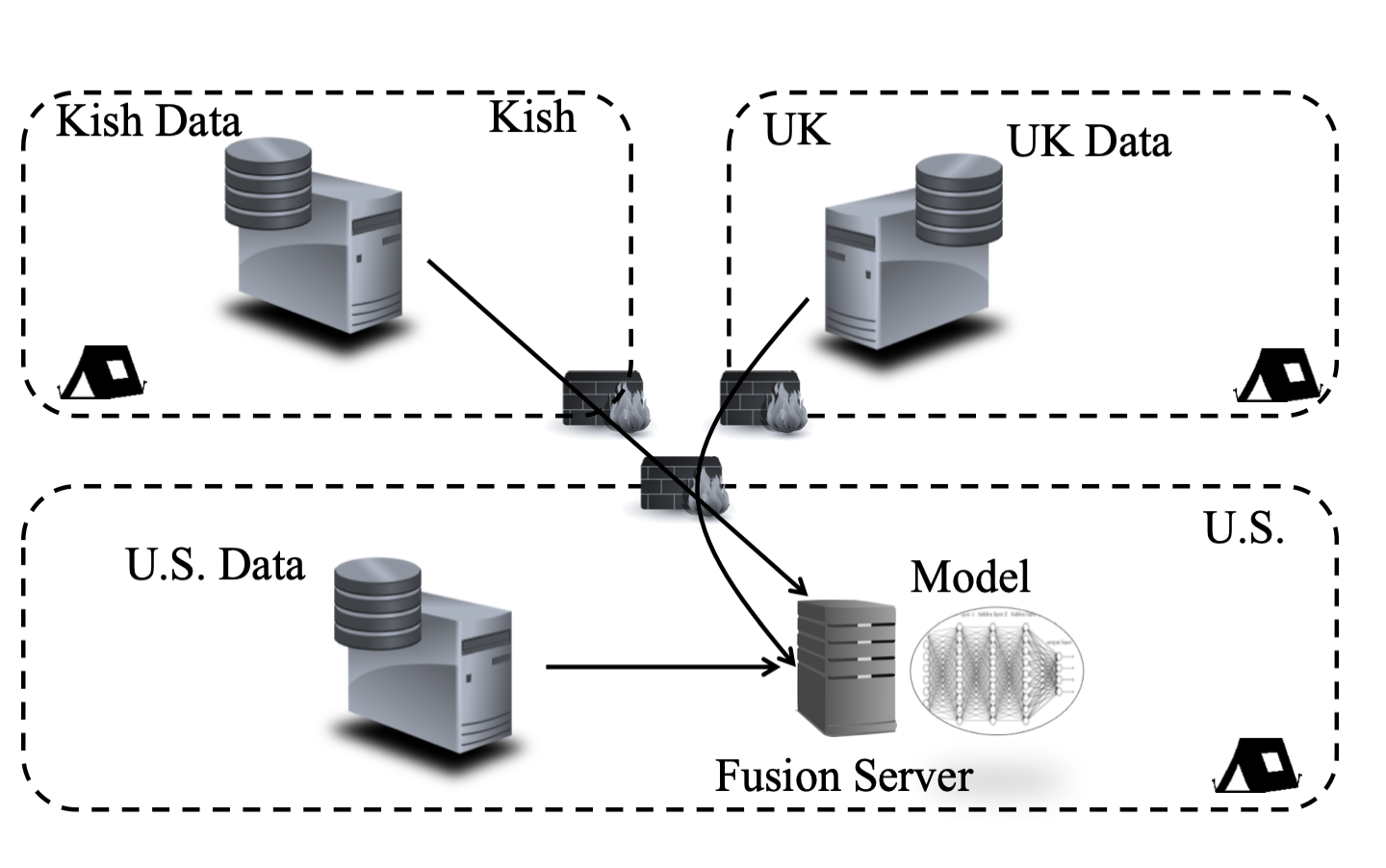}
      \caption{The Data Sharing Mode for Federated Learning}
      \label{fig:fig2}
\end{figure}

\textbf{Data Sharing Mode:} In the data sharing mode shown in Fig.~\ref{fig:fig2}, each coalition partner has a set of training data that they may have obtained from their individual surveillance or other data collection operations. The partners are willing to share the training data with each other so that they can build a better model. One or more of the partners who have adequate computing resources to train an AI model would receive the data from all of the other partners, and create a model from the consolidated data. The data may be sent using data summarization techniques like coresets~\cite{lu2019robust} or sketches~\cite{ko2019data}. 

Such a mode for federation during the learning phase is shown in Figure~\ref{fig:fig2} where three coalition partners, Kish, UK and U.S. are using the computational power available to the U.S. to create a common model. 

\begin{figure}[h]
      \centering
      \includegraphics[width=0.8\columnwidth]{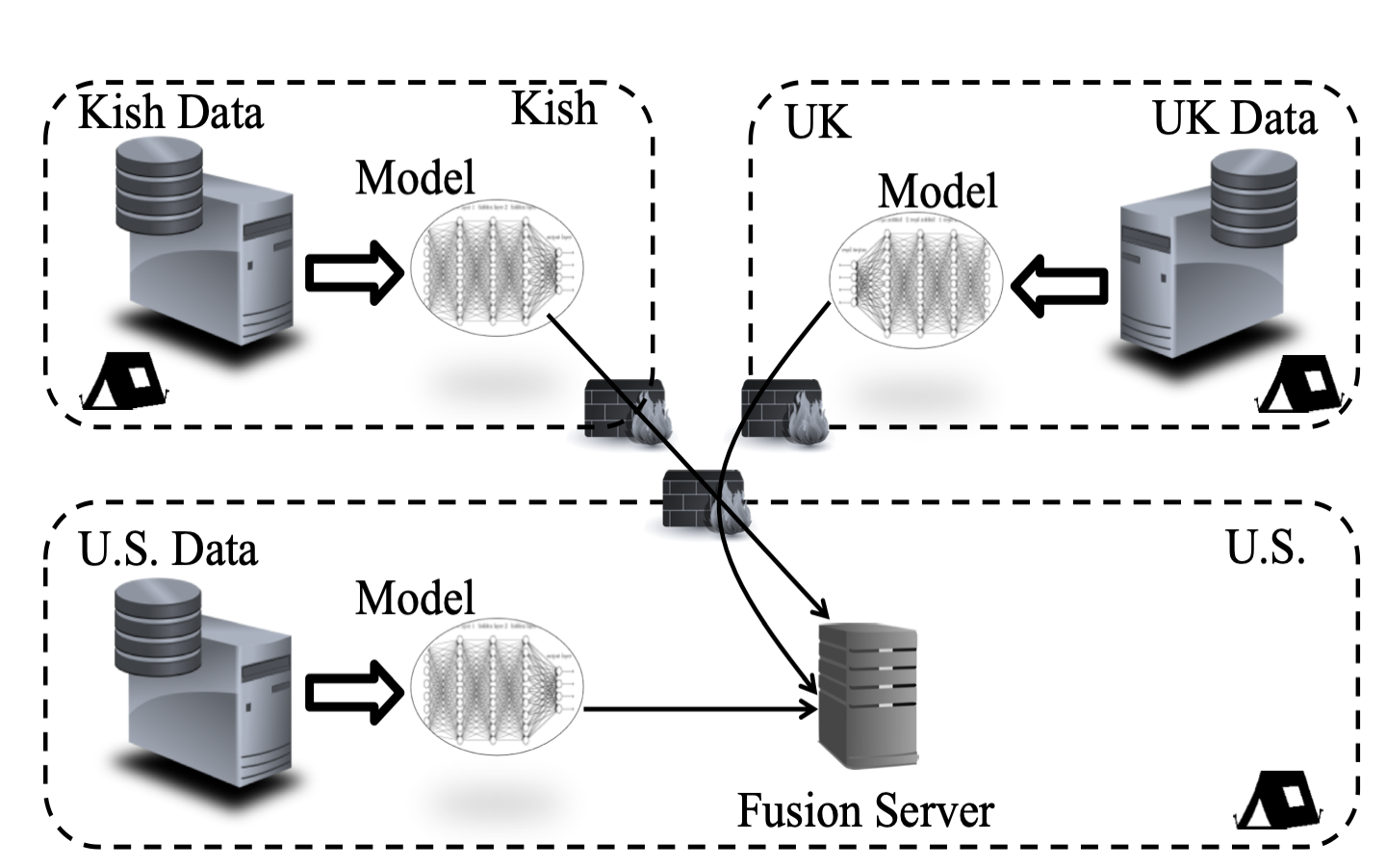}
      \caption{The Model Sharing Mode for Federated Learning}
      \label{fig:fig3}
\end{figure}

\textbf{Model Sharing Mode:}
In the model sharing mode shown in Fig.~\ref{fig:fig3}, the coalition partners do not share the training data with each other. Instead, they each train their models on the local data they have, and exchange models with each other. This mode is useful when the training data sets are too large, or the training data sets can not be exchanged for any reason, e.g. they may reveal sensitive details about the attributes of the equipment used to collect the data. Model sharing may use approaches for fusion of models ~\cite{verma2018algorithm} which can be optimized for performance ~\cite{wang2018edge}.  

Such a mode for federation during the learning phase is shown in 
Figure~\ref{fig:fig3} where three coalition partners, Kish, UK and U.S. are using a fusion server hosted in the U.S. to create a common model.
The advantage of such sharing is the ability to capture different patterns available in the training data present in different coalition members.

\section{Challenges in Federated Learning}
\label{sec:challenges}

Several technical challenges need to be overcome in order for coalition partners to share their data or model during the learning phase of federated learning. In order to deal with them in a consistent manner, giving the wide diversity of AI models and their applicability domains, we can divide any AI enabled task into the four categories shown in Table~\ref{tab:tab1} based on the input and output used by the task. A more detailed justification of this categorization is found in~\cite{enterprise2019}.

\begin{table}[h]
\centering
\caption{Categories of AI tasks}\label{tab:tab1}
\begin{tabular}{|c |c |c |} 
\hline
\textbf{Category} & \textbf{Input} &  \textbf{Output} \\ 
\hline\hline
I &  Features & Known \\ 
\hline 
II &  Raw & Known \\ 
\hline
III & Features & Unknown \\
\hline 
IV & Raw & Unknown \\
\hline
\end{tabular}
\end{table}

Based on the above categories and the two modes for coalition federated learning, we can identify the following challenges that need to be overcome for both modes of operation: 

\noindent\textbf{Varying Data Quality:} For training data that consists of raw input (categories II and IV in Table~\ref{tab:tab1}), the input modality could vary from site to site.  For example, a coalition partner may be storing their input from surveillance cameras in a low-resolution format, while another may be storing the input in a high-resolution format. 
Combining models or data based on different modalities may be problematic, but important. \\
\noindent\textbf{Linguistic and Lexicon Mismatch:} 
For training data that consists of known output (categories I and II in Table~\ref{tab:tab1}), the labels that are assigned to the outputs at different sites may differ; similarly, training data that consists of features (categories I and III in Table~\ref{tab:tab1}), the names of the features that are used at different sites could also be different.  This is due to the varying terminologies used at different locations, thus resulting in ambiguous and conflicting labeling conventions.

For the model sharing mode, we have the following additional challenges: 

\noindent\textbf{Incomplete Knowledge:} Data at each site may only contain a fragment of the whole data landscape.  Thus, there could be a missing \textit{feature problem}, i.e., for training data that consists of features (categories I and III in Table~\ref{tab:tab1}), some features may only be present at some of the sites.  Moreover, for training data that consists of known output (categories I and II in Table~\ref{tab:tab1}), some of the labels may be missing at some of the sites. We refer to this as \textit{missing classes} problem and this can have a significant impact on the accuracy of federated learning~\cite{juliervermacirin2018}.  Finally, there may be situations in which some \textit{values are missing} from the underlying schema, e.g., training data that consists of featured input (categories I and III in Table~\ref{tab:tab1}), if missing values are found across sites it must be ensured that these are resolved in the same way at each site.\\
\noindent\textbf{Skewed Data}: For training data that consists of known output (categories I and II in Table~\ref{tab:tab1}), the distribution of classes in some sites may be entirely different from the distribution of the classes in other sites. For training data that consists of features (categories I and III in Table~\ref{tab:tab1}), the data collected at different sites may correspond to different regions of the feature space. Extrapolating across different regions of applicability can cause mistakes in the estimation of the AI model~\cite{dataskew2019}. Different sites may also have collected different volumes of training data, which may make some trained models to be better than others.\\
\noindent\textbf{Synchronization}: Depending on the level of coordination among different sites, they may be able to conduct their training in synchronized manner with each other. In other cases, such synchronization may not be possible and federation may need to happen in an asynchronous manner.  Asynchronous operation may particularly be required in cases where one site has more/less processing power or capability than another site to avoid one partner unduly having to wait for the other to complete each round of training.

The basic approach we use to solve the challenges in federated learning is to have a mechanism which can generate policies that address the challenges in federation of both training data and models, both provided by different coalition partners. In the next section, we describe such an approach.  
\section{Generative Policies}\label{sec:gen-policy}
In this section, we discuss the architecture for generating policies. In the rest of the paper, we describe the type of policies that need to be generated, and the template such policies need to follow. 

In order to generate their own policies, the different services in the system, e.g. the different components responsible for training models, sharing data, or for evaluating the quality metrics of a data or model, need to generate their own policies. They generate the policies in order to meet the objectives that are provided to them by humans as guidance. 

\begin{figure}[h]
      \centering
      \includegraphics[width=0.7\columnwidth]{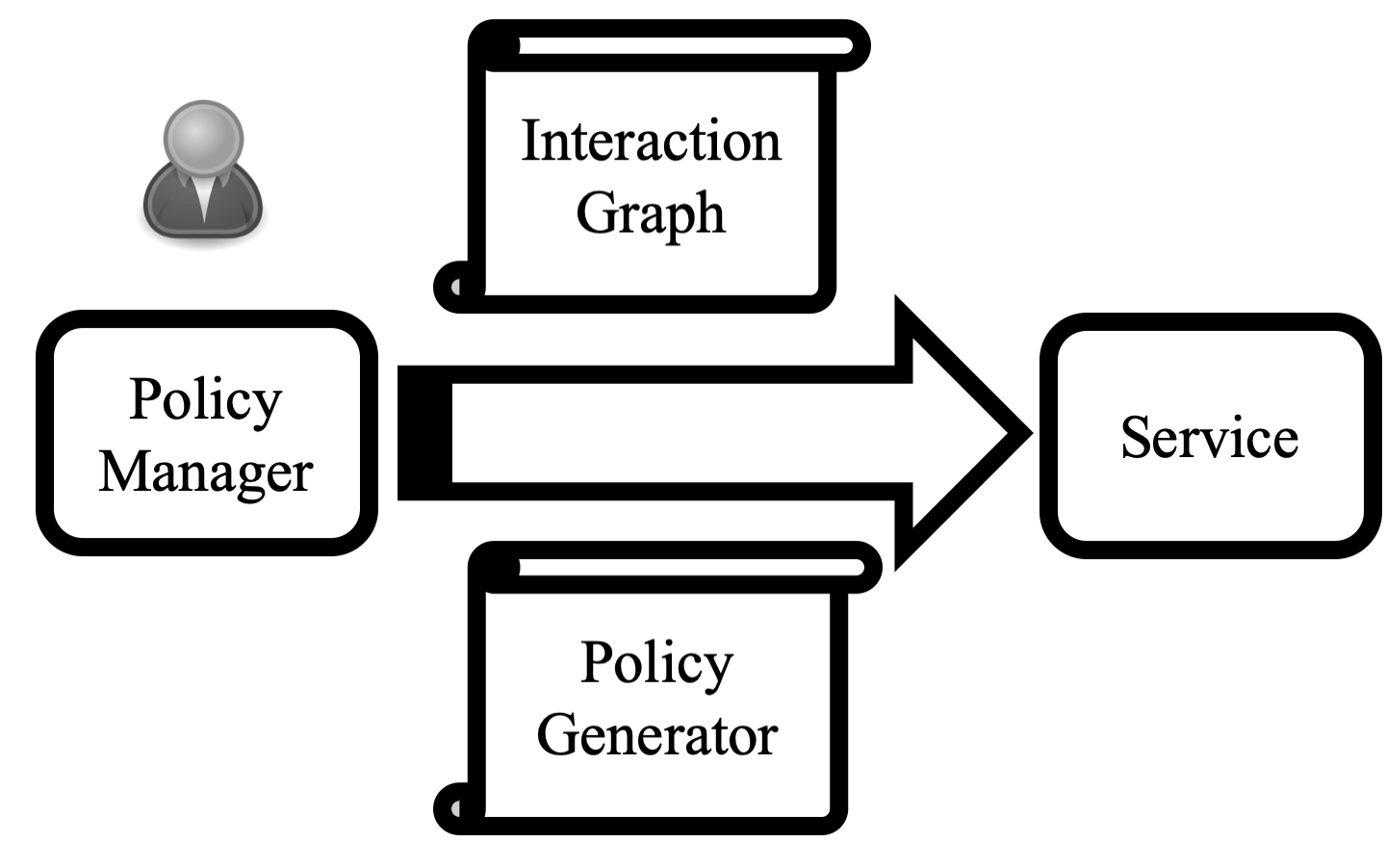}
      \caption{Policy Generation Architecture for Services and Devices}
      \label{fig:figp-1}
\end{figure}

In the overall approach, shown in Figure~\ref{fig:figp-1}, the human guidance usually consists of two pieces of information, a definition of context and a policy generator. The policy generator may be either an Answer Set Grammar~\cite{caloagenp,white2019smartcomp}, or a template to generate policies, or any other mechanism to create the policies. The context would usually consist of an interaction graph that defines which types of other services or devices a specific device/service could expect to see in the environment, and the attributes one would expect of each of those types of systems. 

The policy generator specifies a policy generation language that is defined over the attributes of the devices/services found in the environment. 


\begin{figure}[h]
      \centering
      \includegraphics[width=0.6\columnwidth]{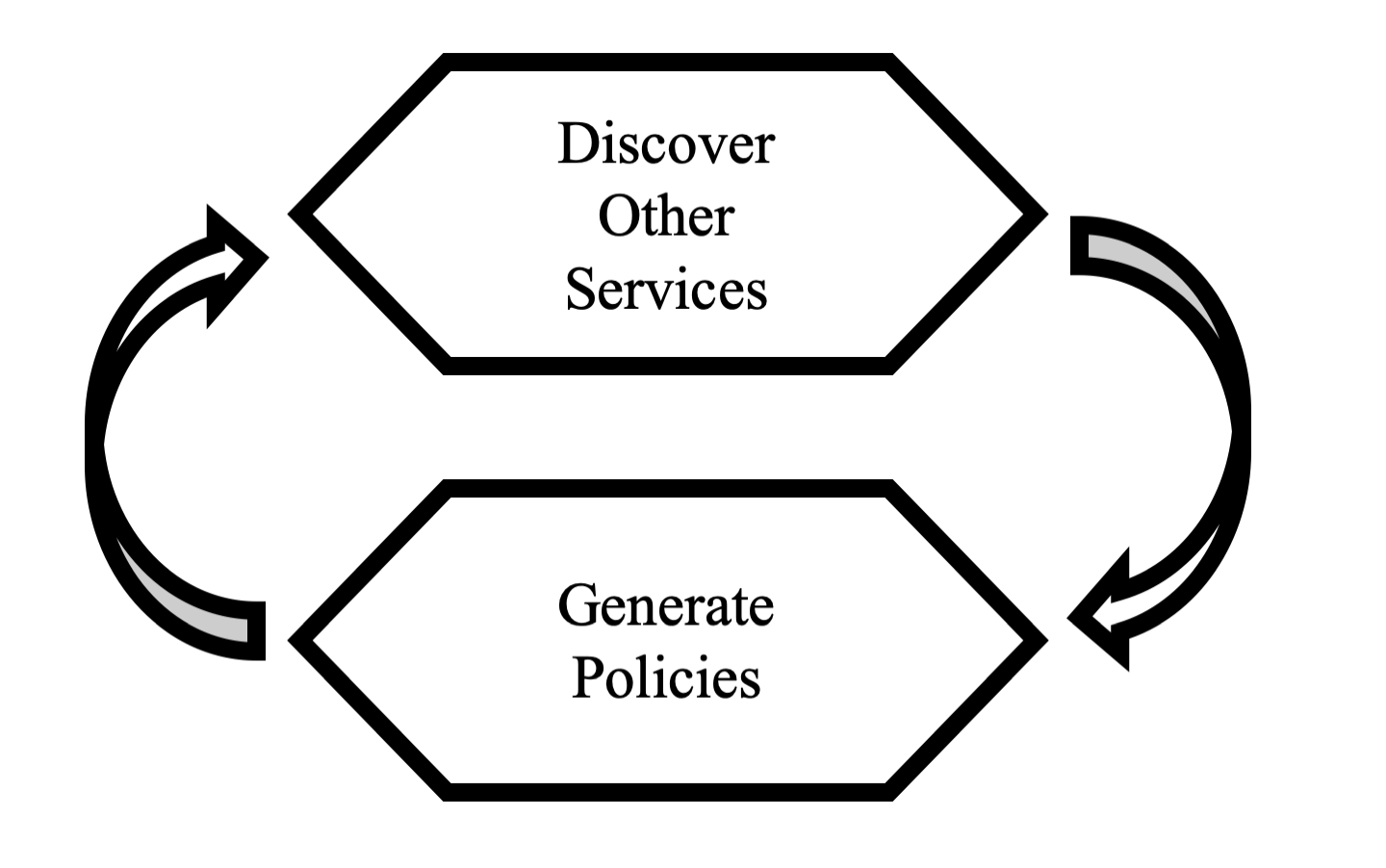}
      \caption{Policy Generation Architecture for Services and Devices}
      \label{fig:figp-2}
\end{figure}

The action of devices is shown in Figure~\ref{fig:figp-2}. When the device or service receives human guidance, it looks for situations where the context changes, and a new policy needs to be generated. A typical instance for security management will be when the set of devices and services in the system   change, e.g. a device/service enters or leaves the system, or changes its role within the system. In those cases, the device/service uses the generator to find the right policies. 

Specifically in the case of federation during the learning phase, context may change when new data or a new model is obtained from a coalition partner, a new coalition partner joins the ecosystem, or a new auxiliary service that can perform tasks like assessing the quality of the offered data or offered model is obtained.  In situations where the context change is beyond those previously seen in any of the training samples then new representative training data may need to be acquired such that a revised generative policy model may be learned.

Given the challenges that need to be overcome for federated learning, it is worthwhile discussing the benefits and trade-offs at a higher level of abstraction. In the next section, we make those observations.  
\section{Observations on Federated Learning}
\label{sec:obsrv}

A key benefit of federation in the learning phase is that it increases the amount of training data available to create a model, and a drawback is that coalition partners may not be trusted to have data that is maintained with sufficient quality, or may not have sufficient value. 

\subsection{Training Data Size}

One of the obvious benefits of federation is that it results in a model that is trained using a larger corpus of training data. Therefore, an important aspect of federation during the learning stage is whether it results in an improvement in the quality of the model that will be created by sharing the content. At the same time, accepting the training data from a partner who may be collecting data using different modalities may increase additional noise in the domain and we must be protected against the adversarial introduction of poor quality training data.

An examination of the training data that would be required for a machine learning problem was done in~\cite{cirincione2018security} in the context of learning a mapping from a set of D elements to a set of R elements, with $\nu$ denoting the noise parameter - probability of observing a non-existing mapping. The AI learning process was mapped to the coupon collector's problem and it was shown that the precision $p$ of an AI model with a training data of size $Q$ can be approximated as:
\begin{center}
    $p = 1-C_0 {e}^{-2Q (1-\nu) } $,
\end{center}
where $C_0$ is a constant depending on the domain of AI model. Similarly, the recall  $\epsilon$, of the AI model would be bounded by:
\begin{center}
   $ \epsilon <= 1 - C_1 {e}^{-Q (1-\nu )}. $
\end{center}
Obtaining the training data from other partners results in an increase in the size $Q$, which should improve both precision and recall. At the same time, it may also potentially increase the noise in the training data, which would have an adverse impact on precision and recall. 

The effective increase in the size of the training data set due to obtaining data sets from other partners would depend on the overlap between the data points collected by different coalition partners. Suppose there are N partners, the $i^{th}$ of which has collected a data set of size $Q_i$, and the noise parameter in the collection process of the $i^{th}$ partner is $\nu_i$. Suppose the resulting training data set after removing duplicates is of
size ${k_i}{Q_i}$ (i.e. $k_i$ times the data set of the $i^{th}$ partner, where $k_i$ is a number between 1 and $\sum_{j}{Q_j}/Q_i$, depending on the duplication among the different data sets. The noise parameter for the resulting data set would be at most $\nu_{agg}$ = $\sum_j{\nu_j}.{Q_j}/{k_i Q_i}$.   
The net resulting precision and accuracy bounds of the overall data set can be estimated from the formula given above. 

As long as $k_i (1-\nu_{agg}) > (1-\nu_i)$, the performance metric of the $i^{th}$ coalition partner will  better by accepting the training data from the other partners. 

An alternative way is to consider the new training data set after sharing to be $Q_{n}$ with a noise rate of $\nu_{n}$, and the original data set to be of size $Q_{o}$ with a noise rate of $\nu_{o}$. The partner is better off with sharing data if $Q_{n}(1-\nu_{n}) > Q_{o}(1-\nu_{o}) $, which  is equivalent to 
\begin{center}
    $Q_{n} - Q_{o} > Q_{n}.\nu_{n} - Q_{o}.\nu_{o} $,
\end{center}

Since the difference marks the amount of new data that is free of noise, it is better for coalition partners to exchange data as long as they are getting new noise-free data points. Note that since the  bounds  ~\cite{cirincione2018security} assumed that the noise rate is small, this observation is valid as long as the partners have data with small amount of noise. 

The effective increase in the training data set due to sharing of data or the model would depend on the independence among the training data that is collected. If the different data sets are collected completely independently, the training data size would be summed up as more data is obtained. However, if the same good data points are being collected, then we may not get a benefit from exchange of data, if the new points are all noisy. 

\subsection{Quality and Value of Information}

The concerns with noise present in partner data, and the concept of independence of data sets can be viewed as rudimentary estimates of the quality and value of information (VoI) present at different coalition partners. In a high level sense, quality of information (QoI) refers to the intrinsic attributes of a piece of data (e.g. how much error or noise the data has) while value of information measures the usefulness of a piece of information that is offered by by coalition partner for a specific purpose, e.g. in improving the precision or recall of an AI model using that additional piece of information. 

In a more general view of quality and value of information, we can assume that there is a ground truth in an environment under consideration which is collected into a training data set by each of the partners. This training data set is converted to a model by the training process. The ground truth and the collected training data set can be viewed as representing points in an information space (\textit{IS}). The model represents an output or a decision, which can be viewed as belonging to a decision space. 

We assume that a distance metric $\Delta$ is defined over the information space, so that \textit{$\Delta$(f,g)} measures the distance between any two elements of the information space. Similarly, a distance metric $\delta$ is defined in the decision space to measure the distance between two decisions. 

The analysis task (model training in this case) is a mapping from the information space to the decision space. Denoting the decision corresponding to an existing information \textit{I} as \textit{a(I)}, the VoI for a new information \textit{J} for an analysis task denoted as \textit{A} which is already using a given information \textit{I}, would be given by 

\begin{center}
\textit{VoI(J $|$ I,A)=$\delta$(a(J+I),a(I)) }
\end{center}
Note that VoI depends both upon the analysis task, as well as the available information that is used for analysis. 

Given any piece of training data set \textit{I} (\textit{I} $\in$  \textit{IS}), which is supposed to be representing a ground truth \textit{G} (\textit{G} $\in$ \textit{IS}), the QoI of \textit{I} is given by 
\begin{center}
QoI(\textit{I}) = $\Delta$($I$,$G$)
\end{center}
where $\Delta$ measures the distance between two points in the information space. 

In contrast to VoI, measures of QoI are dependent only on the attributes of the collected data and the properties of the collection process, and independent of the analysis task or the amount of current information available to the system. 

The concepts of QoI and VoI of training data sets and models provide us a way to deal with the various challenges associated with accepting data from coalition partners. A partner accepts the data or model offered by a partner only if the offered data/model has a high quality of information and a high value of information. In other cases, the model is rejected. Policies determining the QoI and VoI thresholds can be defined and used to influence the operation of federated learning in both modes. 

Given these observations, in the next two sections, we envision a data curator based on the generative policy model described in Section~\ref{sec:gen-policy}. 
\section{Federated Learning in Data Sharing Mode}
\label{sec:data-sharing}

As mentioned above, understanding the quality and value of data offered by different coalition partners provides a way to determine whether or not to access the data to build a common model. One approach to create an improved model is to have a data curator system that would be able to examine these attributes as training data is received from the different partners, and then to generate policies which are able to reject data that does not offer a good quality or a good value. 

The data curation system can be viewed logically as a data aggregator and filter that takes the training data provided by all of the coalition partners, and combines them into a single training data set. 

Such a system to create a data curator that can generate its own policies is described in~\cite{padg2018gen} and~\cite{verma2019managing}. The data curator relies on a series of helper services to determine whether or not to accept any data that is offered by the partner. These helper services include services that can convert formats of data into various modes, and determine the QoI and VoI of an offered training data set. The QoI is obtained by examining different attributes, such as the consistency of the offered data against the data that is already present at the curator and accepted as safe, the balance or imbalance in the classes available from different training data that would be received, and increase in the size of independent data sample points. 

Let us examine the approaches by which this data curator would deal with the different challenges of federation that were identified in Section~\ref{sec:challenges}

\noindent\textbf{Varying Data Quality:} The difference in data quality among different sites can reflect itself as a difference in formats, a difference in resolution of the input, or a difference in the level of trust that may exist in accepting data from the remote partner.

In order to address the format mismatch issues, the data curator would need to determine whether the format of the data provided by the coalition partner can be translated to the local format that is used by the training algorithm. The curator does this task by generating format translation policies, which automatically convert any data received from a partner into a canonical format. These policies take the format of: 

\begin{center}
\footnotesize{
\pseudocode{
 if (source-name == XXX) and (source-format = YYY) then invoke helper-service ZZZ.}}
\end{center} 

In this particular case, any data in format YYY coming from the coalition partner XXX is first converted into the canonical format by calling the helper service ZZZ. 

Another type of acceptance policy needs to be generated which takes the format of: 
\begin{center}
\footnotesize{
\pseudocode{ if (source-name == XYZ) and (label == L1) then accept/reject data.   }}
\end{center}

Alternatively, if trustworthiness values have been assigned to each of the coalition partners, the following policies need to be generated. 
\begin{center}
\footnotesize{
\pseudocode{ if (source trustworthiness $>$ threshold) and (label == L1) then accept/reject data. }}  
\end{center}

When data is being analyzed for QoI and VoI, acceptance policies of the following format need to be generated. 
\begin{center}
\footnotesize{
\pseudocode{ if (data QoI $>$ threshold) and (data VoI $>$ threshold) then accept/reject data.   }}
\end{center}

\noindent\textbf{Linguistic and Lexicon Mismatch:} 
In order to deal with differently named labels and features, the data curator needs to generate the relabeling policies that would convert all the incoming labels provided by the coalition partner into a single canonical set of labels. In order to do so, the data curator needs to generate data relabeling policies, which will take the format:

\begin{center}
\footnotesize{
\pseudocode{if (source-name == XXX) and (feature-name = YYY) then change label to ZZZ.}}

\footnotesize{
\pseudocode{if (source-name == XXX) and (field-name = YYY) and (label-name = ZZZ) then change label to QQQ.}}
 \end{center}

Because of the fact that data is being collected in a single location, many of the other issues, such as skew in data, missing labels etc. are resolved just because they are being done at a central site. 

The algorithms and system which can generate the policies of the type described above automatically are described in~\cite{padg2018gen} and~\cite{verma2019managing}.
\section{Federated Learning in Model Sharing Mode}
\label{sec:model-sharing}

\begin{figure}
      \centering
      \includegraphics[width=0.7\columnwidth]{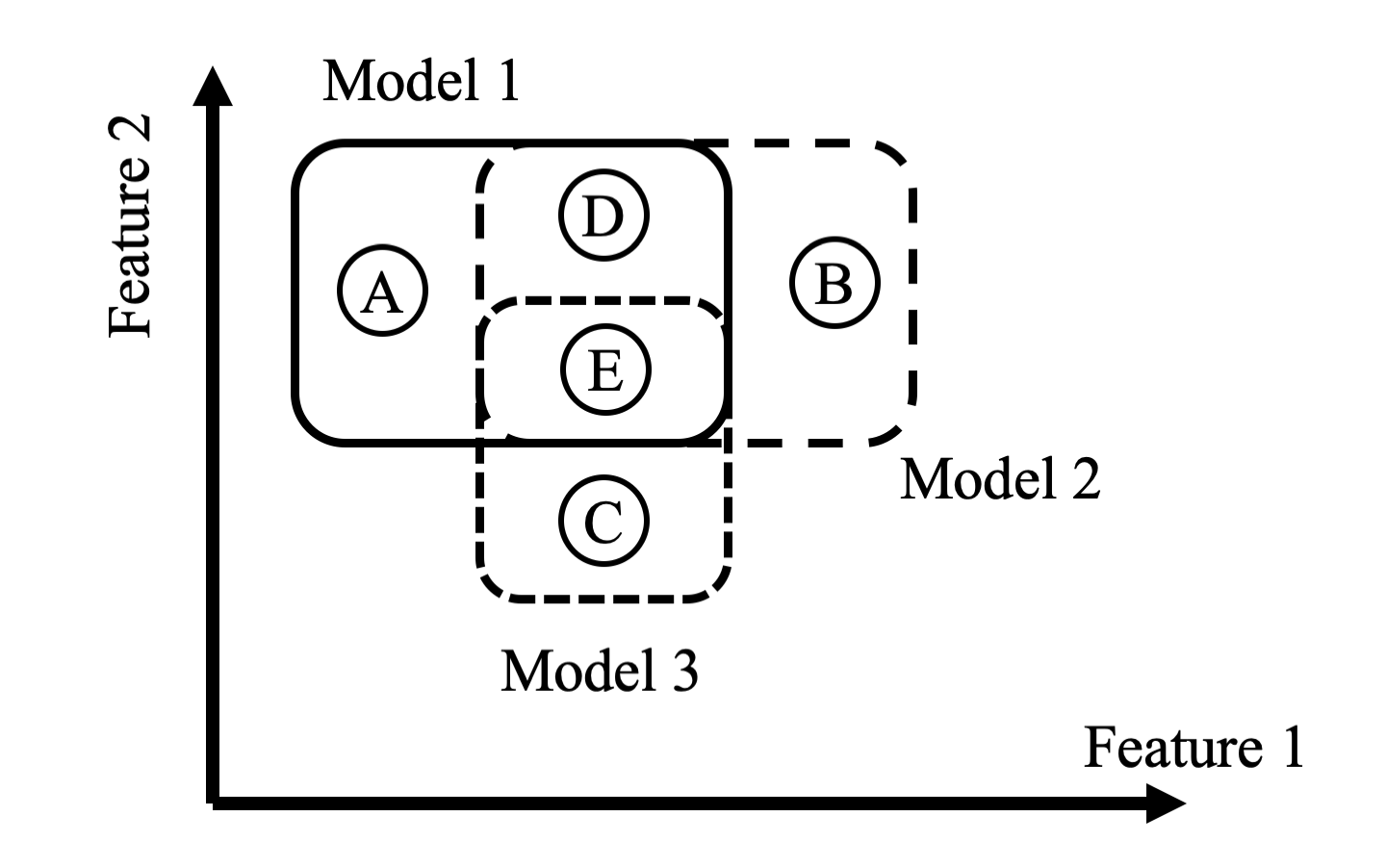}
      \caption{Ensemble Approach for Federation }
      \label{fig:fig4}
\end{figure}

In model sharing mode, different models are received from other coalition partners, and they are combined together in order to make a composite model. In addition to the task of converting models to a canonical format by generating policies for format conversion, and quality based acceptance of models, we need to consider how to address the challenges arising out of incomplete knowledge, skewed classes, size differences and synchronization. 
The typical federation algorithm for combining models in the current state of the art requires iterative averaging of the weights of models conforming to the same architecture~\cite{mcmahan2016communication, bonawitz2016practical}. However, this iterative addition requires operation in a synchronized manner, and a common model architecture. 

In order to address synchronization issues, different flavors of federation can be used. We have proposed two different flavors for model fusion algorithms~\cite{juliervermacirin2018}, the first one is used when different sites can coordinate their training processes, and the second one is used when the different sites need to conduct their training processes in an unsynchronized manner. In the former, the different sites perform averaging of a common architecture after training on mini-batches. In the latter, the different sites perform model training over all of their data, and averaging the model weights after the model has been fully trained in a round-robin fashion. The round-robin approach converges after some number of rounds are completed. 

\begin{figure}
      \centering
      \includegraphics[width=0.7\columnwidth]{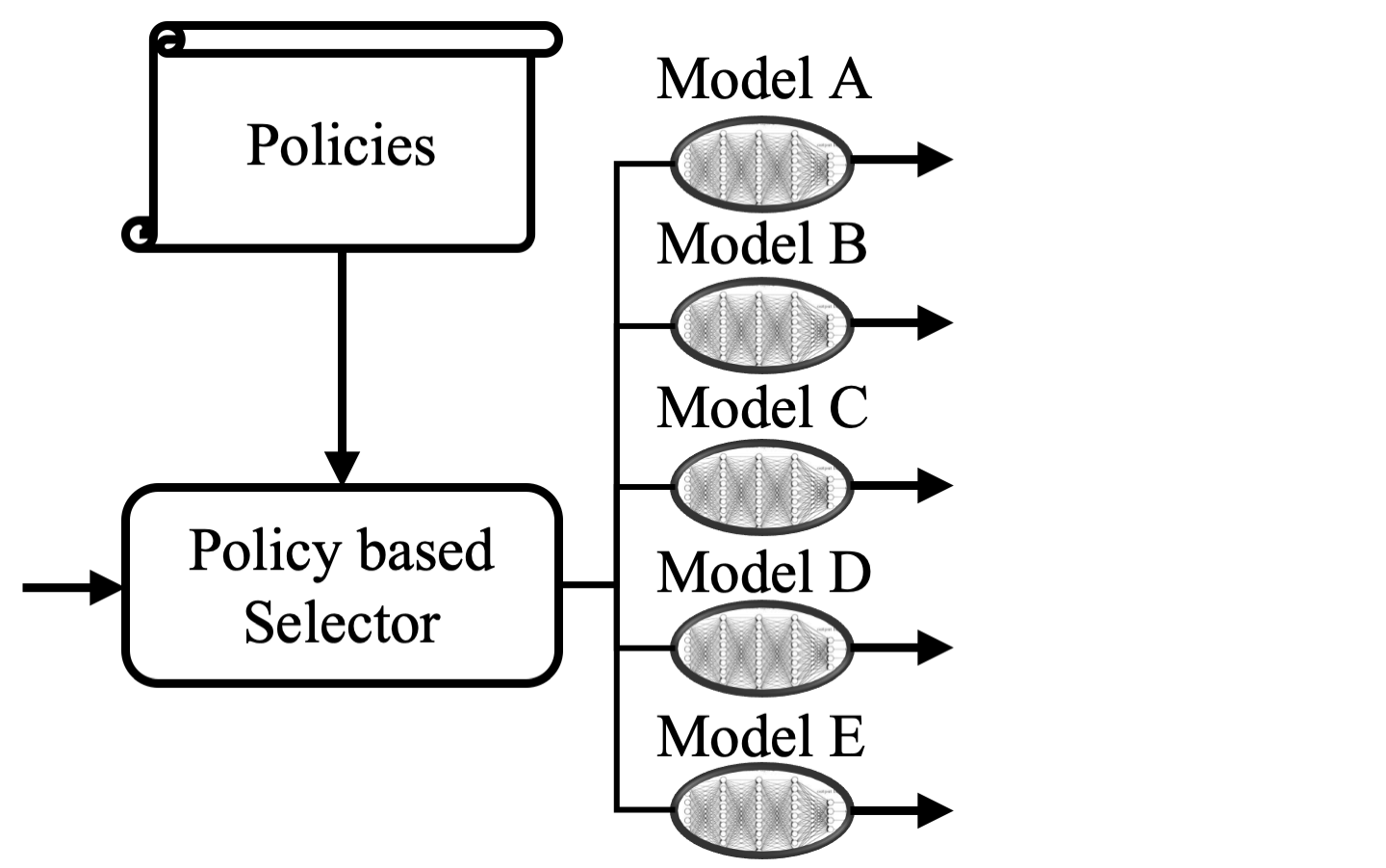}
      \caption{Inference using Ensemble}
      \label{fig:fig5}
\end{figure}

In order to address the challenges associated with skewed classes and skewed features, we have proposed two approaches. The first approach~\cite{juliervermacirin2018} is to consider the exchange of a small number of samples to counter the skewness of classes.  The second approach~\cite{dataskew2019} examines the properties of the data that is used for training at each site, and determines what the applicability boundaries for the use of the model at each site is. Model fusion is done only for those regions of data where the training data at each of the sites is applicable. The net result is an ensemble of models along with a set of criteria to check which of the many models to use for a given piece of inference.

The second approach is illustrated visually in Figure.~\ref{fig:fig4}. In the figure, three models, each provided by a different coalition partner are shown when plotted in the space of two features. These features could be the two principal components of a multi-variate data matrix determined using principal component analysis~\cite{wold1987principal}. In the space defined by the two axes shown in the figure, we can identify the regions shown as show in Table~\ref{tab:tab2}. 

\begin{table}
\centering
\caption{Regions and Applicable Tasks}\label{tab:tab2}
\begin{tabular}{|c |c |c |} 
\hline
\textbf{Region} & \textbf{Applicable} &  \textbf{Federation} \\ 
\textbf{} & \textbf{Models} &  \textbf{Approach} \\ 
\hline\hline
A &  1 only & Use 1 only \\ 
\hline 
B &  2 only & Use 2 only \\ 
\hline
C & 3 only & Use 3 only \\
\hline 
D & 1 and 2 only & Federate 1 \& 2 \\
\hline
E & 1, 2 and 3 & Federate 1, 2 \& 3 \\
\hline
\end{tabular}
\end{table}

The ensemble approach would create 5 models, the original 3 models provided by the individual coalition partners, a fourth model obtained by federating models 1 and 2, and a fifth model used by federating all three models together. This ensemble is front-ended by a policy-based model selector, which is driven by policies which select the right model depending on where the input is in the desired feature space. The structure of the ensemble would be as shown in Figure~\ref{fig:fig5}. The federation process would determine the policies for the policy based selector which will be used to choose among the different federated models. 

The generic format of this policy would be: 

\begin{center}
\footnotesize{
\pseudocode{ if (component1 $\leq$ threshold1) and (component1 $\geq$ threshold2) and (component2  $\leq$   threshold3) and (component2 $\geq$ threshold4) then use model model-no.   }}
\end{center}

In this particular case, the policies will be generated so as to correspond to the region boundaries defined in Table~\ref{tab:tab2}

The benefit of the second approach is that it can be used in conditions where coalition partners are unable to exchange any amount of data. However, it can only be used when regions of applicability of training data can be clearly identified, e.g. when the AI tasks belongs to Category I or III as defined in Table 1. The benefit of the first approach of data exchange is that it is applicable to all four categories of AI tasks as defined in Table I. 

\begin{figure}[h]
      \centering
      \includegraphics{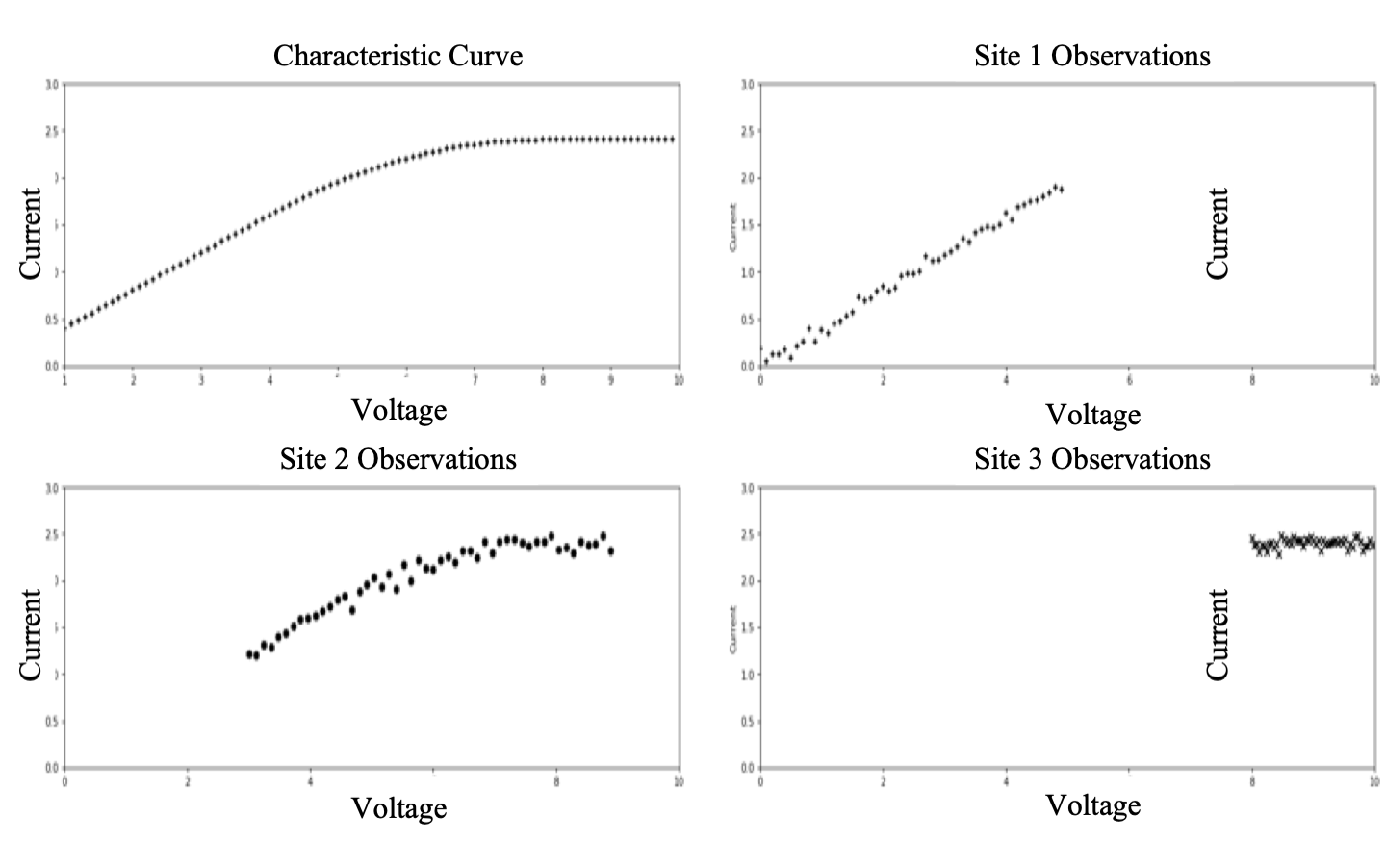}
      \caption{Ground Truth and Individual Site Observations}
      \label{fig:fig6}
\end{figure}

\begin{figure}[h]
      \centering
      \includegraphics{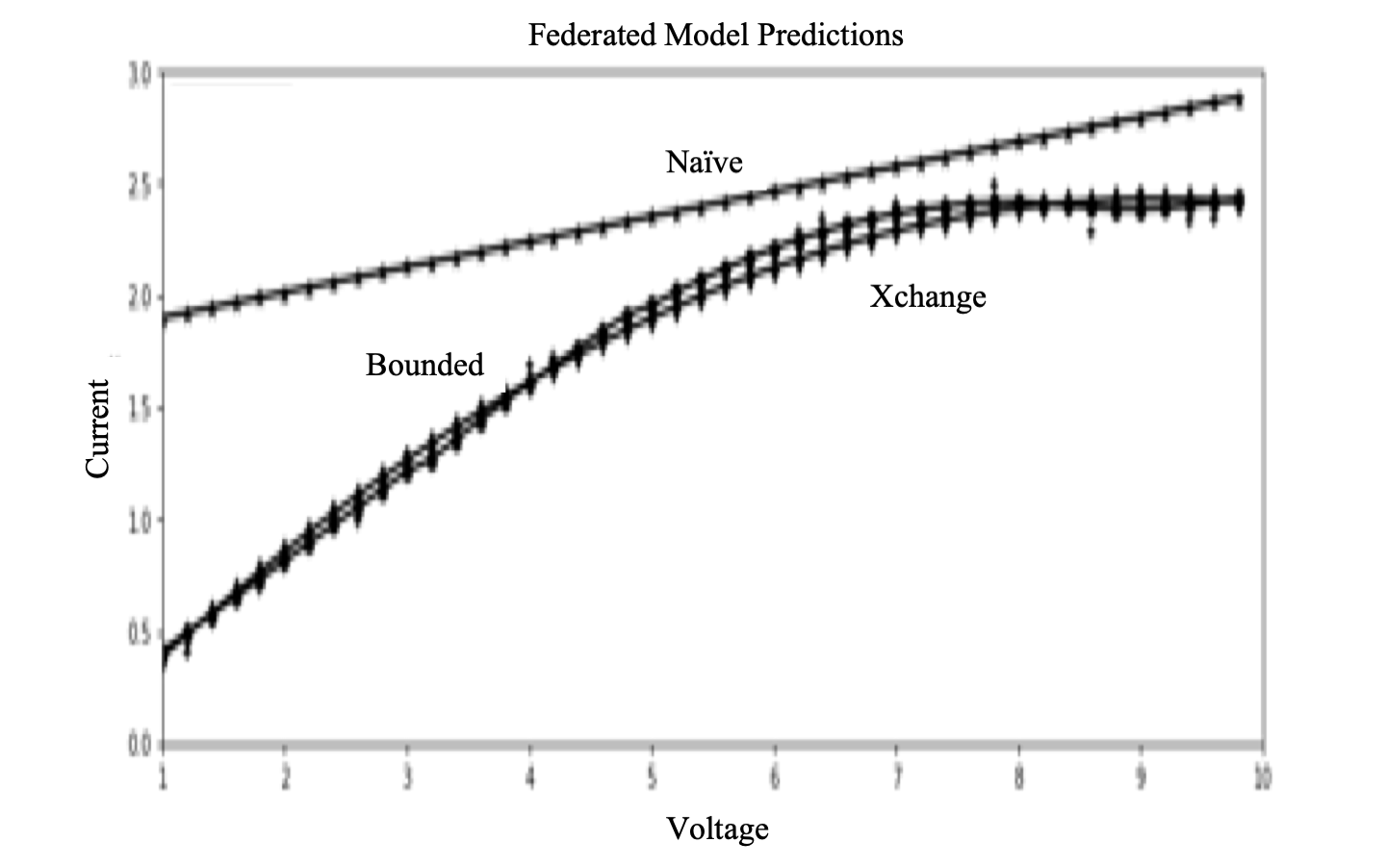}
      \caption{Performance of Federation Approaches}
      \label{fig:fig7}
\end{figure}

A simple illustration of the effectiveness of the two approaches in improving the quality of federated learning in presence of skewed data is shown in Figures~\ref{fig:fig6} and~\ref{fig:fig7}. 

Figure~\ref{fig:fig6} shows the characteristic curve (i.e. the ground truth) which needs to be determined for a specific data set, e.g. modeling an electronic component whose attributes like voltage and current are measured independently at three different locations. The three locations examine the component in different conditions, which results in the three different relations being learnt by each of the sites, if they do not share their information. None of the three relationships captures the ground truth completely. 

Figure~\ref{fig:fig7} shows 
the model that is returned
when fusion of model parameters is done in a naive manner without taking into account the applicability of the models. The performance of both approaches, by considering the bounds and by exchanging a portion of the information to extend the region of applicability, is shown in this figure. While the naive model does not provide a good approximation to the ground truth shown in Figure~\ref{fig:fig6}, the two approaches perform a reasonably good approximation of the same. 

\subsection{System Implementation}

\begin{figure}[thpb]
      \centering
      \includegraphics[scale=0.8]{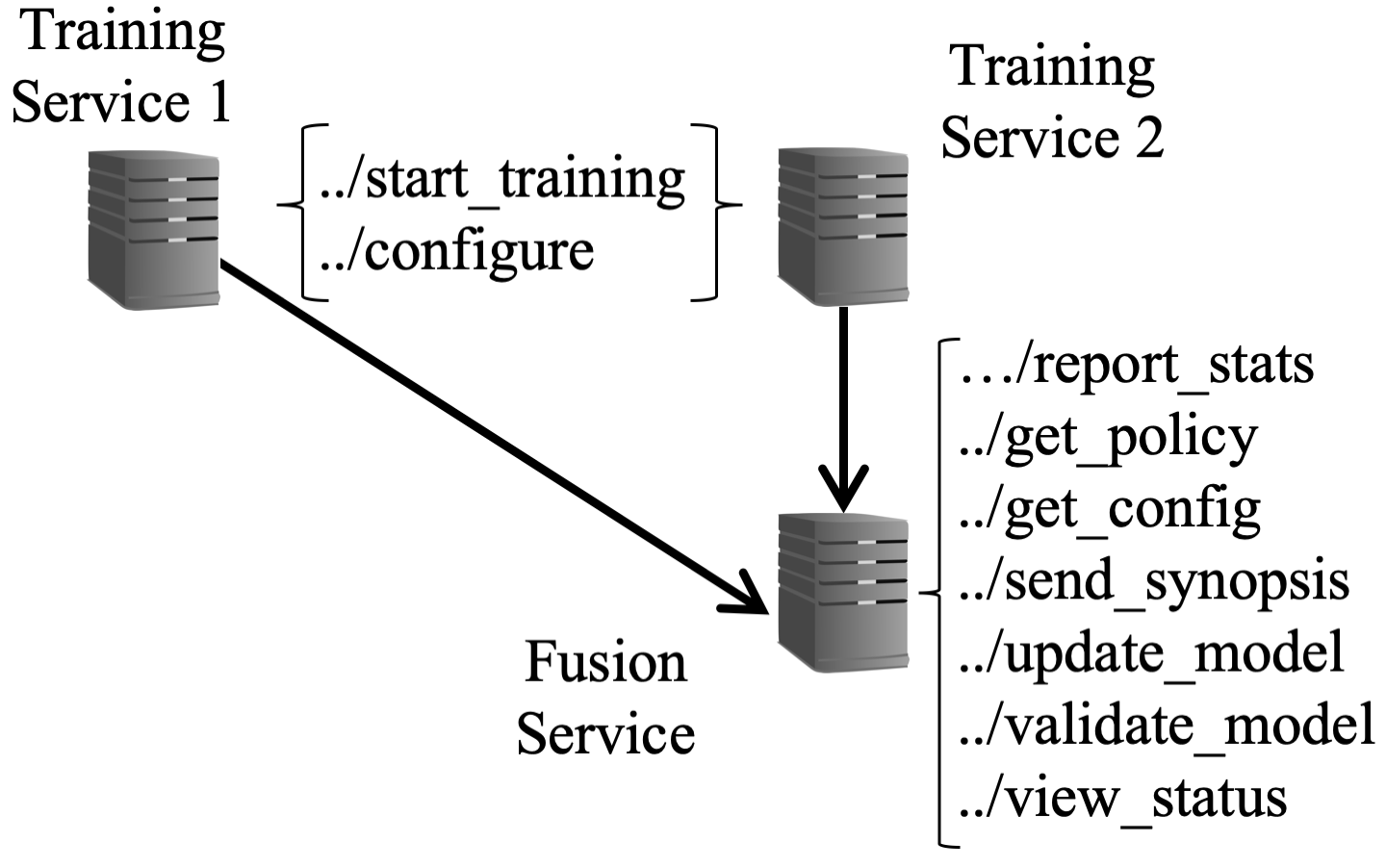}
      \caption{The set of services implementing Model Fusion}
      \label{fig:fig10}
\end{figure}

All of the components described above in this section can be implemented together into an integrated system that can provide a single model fusion service for several partners. From an overall system architecture perspective, this can be viewed as a training service which is implemented by all the coalition partners, and of a fusion service which is implemented by the partner training a model. The training service and the fusion service implement a set of services which perform partial training of the model, followed by a fusion of the same (See Figure~\ref{fig:fig10}). 
\vspace{-0.25cm}
\vspace{-0.25cm}
The training service provides only two service calls, one to configure it, and the other to start the training process at each of the clients. These commands can be invoked manually, or in some cases by the fusion server. The fusion server provides a richer set of commands which are invoked by the different training services.

When the training service is invoked via the configuration command, it analyzes the statistics of the locally available training data. It collects the description of the training data that it has available locally. It then invokes the statistics reporting service at the fusion site. 

As a result of this invocation, the fusion server will create a set of control and configuration parameters  for  the  fusion  AI  training  session for all the coalition partners in the session. These include configuration for obtaining the policies generated to convert formats, and configuration parameters for facilitating a data exchange, and parameters needed for the fusion process.  The fusion service sends this control and configuration information to the requesting training services.

On  receipt of the control information, each training service contacts the fusion service to get a set of policies. The fusion server compares the data it has locally with the provided statistics, and uses it to generate a set of policies for transformation of data at the training service. The goal of these generated policies is to get data from each of the different training services into a common schema. The policies that will be sent to the training service will include instructions for changing the type of raw data (e.g. convert .jpg images into .png, or convert .avi sound files into .wav files etc.), relabeling the features to a common set of names, and relabeling the output label values into a different common set. The algorithms for generating these policies are described in more detail in~\cite{padg2018gen}. 
Upon receipt of the policies, the training service uses the policies to convert the local training data into the common format for fusion.  The previously received control information instructs the training server about the operations it should conduct before starting the fusion, e.g. in some types of fusion processes it may need to send a small sample of its data set or a generator for representative synthetic data. The control information may also contain information about batch-sizes and number of iterations the training service may need in order to conduct a successful model training exercise. With the receipt of the control parameters, the training service goes through the training stage, working with the fusion service in the fusion stage. Once the training has been completed, the fusion server may validate the model and compute its performance metrics. This may also include synchronizing with different training services. 


\section{Conclusions}
\label{sec:conclusion}

We have presented a summary of an approach for federated learning among different partners in a coalition during the model training stage of an AI/ML enabled task. We have shown two different modes in which coalition partners can coordinate in this phase, either by sharing training data or by sharing models. We have looked at the different issues that may arise in both modes, and have shown how generative policies can be used to address those issues. 

In continuing research, we will extend the approach of federation and coordination among coalition partners to the other two stages of AI-enabled tasks, effectively enabling federated inference and federated action within a coalition, and examine approaches to maximize the overall effectiveness of an AI-enabled task. There are other issues that are also subjects of ongoing exploration, such as approaches to transform AI models of different architectures into a canonical model. Furthermore, we are also exploring algorithms for policy-based model ensemble when AI tasks are uncertain.  

\balance
\bibliographystyle{IEEEtran}
\bibliography{references}

\end{document}